\documentclass[fleqn,10pt]{wlscirep}

\usepackage[utf8]{inputenc}
\usepackage[cmsy]{MnSymbol}
\usepackage[T1]{fontenc}
\usepackage{textgreek}
\usepackage{mathtools}
\usepackage{stringstrings}
\usepackage{xstring}
\usepackage{xfrac}
\usepackage{tikz}
\usepackage{pgfplots}
\pgfplotsset{compat=1.7}
\usepackage{subcaption}
\usepackage[justification=justified]{caption}

\usepackage{bm}

\usetikzlibrary{trees,arrows.meta,shapes}
\DeclareMathAlphabet\mathbfcal{OMS}{cmsy}{b}{n}
\DeclareMathAlphabet{\mathbfit}{OT1}{cmr}{bx}{it}
\renewcommand{\mathbf}{\textbf}

\newcommand{\event}[1]{
    \overline{#1}
}

\newcommand{\bigCI}{\mathrel{\text{\scalebox{1.07}{$\perp\mkern-10mu\perp$}}}}

\newcommand{\calA}{\mathcal{A}}
\newcommand{\calB}{\mathcal{B}}
\newcommand{\calN}{\mathcal{N}}
\newcommand{\calM}{\mathcal{M}}
\newcommand{\calG}{\mathcal{G}}
\newcommand{\calI}{\mathcal{I}}
\newcommand{\calJ}{\mathcal{J}}
\newcommand{\calV}{\mathcal{V}}
\newcommand{\calD}{\mathcal{D}}
\newcommand{\calH}{\mathcal{H}}
\newcommand{\calX}{\mathcal{X}}
\newcommand{\calY}{\mathcal{Y}}
\newcommand{\calZ}{\mathcal{Z}}
\newcommand{\calF}{\mathcal{F}}
\newcommand{\calP}{\mathcal{P}}
\newcommand{\calW}{\mathcal{W}}
\newcommand{\calbfF}{\mathbfcal{F}}
\newcommand{\calbfX}{\mathbfcal{X}}

\newcommand{\calE}{\mathcal{E}}
\newcommand{\calbfE}{\mathbfcal{E}}
\newcommand{\mathf}{\  \mathit{f}}
\newcommand{\mathbff}{\mathbfit{f}}
\newcommand{\mathbbP}{\mathbb{P}}
\newcommand{\mathbbR}{\mathbb{R}}

\newcommand{\dooeq}{\mathrel{\overset{\mathrm{do}}{\raisebox{-2pt}{=}}}}
\newcommand{\doo}[1]{
 \lowercase{\IfStrEq{#1}}{#1}
    {\text{\,\raisebox{-.25ex}{\textbullet}}#1}
    {\text{\,\textbullet}#1}
}

\DeclareMathOperator{\pa}{PA}
\DeclareMathOperator{\ch}{CH}
\DeclareMathOperator{\an}{AN}
\DeclareMathOperator{\de}{DE}

\DeclareMathOperator{\ap}{AP}

\newlist{enumerate*}{enumerate*}{1}
\setlist[enumerate]{itemsep=0mm}

\title{A Critique on the Interventional Detection of Causal Relationships}

\author[*]{Mehrzad Saremi}
\affil[]{M.Sc. strudent, Department of Artificial Intelligence, Amirkabir University of Technology, Tehran, 1591634311, Iran}

\affil[*]{mehrzad.saremi@aut.ac.ir}

\begin{abstract}
Interventions are of fundamental importance in Pearl's probabilistic causality regime. In this paper, we will inspect how interventions influence the interpretation of causation in causal models in specific situation. To this end, we will introduce \textit{a priori} relationships as non-causal relationships in a causal system. Then, we will proceed to discuss the cases that interventions can lead to spurious causation interpretations. This includes the interventional detection of \textit{a priori} relationships, and cases where the interventional detection of causality forms structural causal models that are not valid in natural situations. We will also discuss other properties of \textit{a priori} relations and SCMs that have \textit{a priori} information in their structural equations.
\end{abstract}
\begin{document}

\flushbottom
\maketitle
%
%
\thispagestyle{empty}

\section{Introduction}
Causal inference is the process of characterizing, detecting, or determining causal relations between physical phenomena \cite{hitchcock1997probabilistic}. There are two major approaches to characterizing causation \cite{hitchcock1997probabilistic, pereira2009modelling}. One is based on the regulatory theory of causality. Within this foundation, two events $A$ and $B$ are thought as cause and effect, if $B$ is regularly observed to follow $A$. As David Hume postulated, within this framework, ``we may define a cause to be an object, followed by another, and where all the objects similar to the first, are followed by objects similar to the second. \cite{hume2016enquiry}'' The other approach is based on probability theory \cite{pearl2009causality}. Advocates of this approach typically use mathematical models called Structural Causal Models (SCMs) as the representative of a causal system. What separates SCMs from ordinary probabilistic models is that they rely on an addendum to probability theory called the ``do-calculus'' \cite{pearl1995causal,pearl2009causality}. The crucial idea in the do-calculus is the notion of interventions. The goal of this paper is a thorough inspection of the applicability of interventions as part of the do-calculus in the determination of causation. We do not intend to discuss the mathematical validity of the do-calculus.

We organize our inspection into two major parts. In the first part, we will investigate the validity of interventions on outcomes with underlying physical proximity or identity. In this part, we will address questions such as ``Is the relationship between the temperature of a star in Kelvin and its temperature in degree Celsius a causal relationship?'', and how interventions respond to such relationships. In the second part, we will discuss whether interventions cause ``unnatural'' behaviour in an SCM. In this part, we want to see whether observing a system under intervention does necessarily infer a model that is valid in the natural setup.

In this paper, we will use measure-theoretic probability. In order to facilitate the inspection of a model --that is probabilistic in essence-- with respect to the world --that is physical--, we will avoid using random variables and use ``outcomes'' as elements that can be both physical and mathematical. We will introduce a ``conceptual taxonomy'' of outcomes, which is more convenient to work with than random variables.

We organize this paper in five sections: In section \ref{sec:basic-notation}, we will introduce the basic notation that we will use throughout the paper. This section includes the basic causality and probability notations, the conceptual taxonomy of outcomes, the graphical representation of SCMs, and finally the definition of SCMs.

In section \ref{sec:background}, we will review a background on the inference of causal relationships. In this section, we will review the problems in the classical detection of causality that have led to the do-calculus. Then, we will define formal interventional rules that are used in the current regime to discover causation directly using experimentation.

In section \ref{sec:distinction}, we will bring about our first inspection. We will make a distinction between two types of relations: those that are known before doing experimentation and those whose inference relies on experimentation. To this end, we will introduce \textit{a priori} and \textit{a posteriori} relations, i.e. the relations that are known prior and posterior to experience. We will argue that \textit{a priori} relations are not causal and that utilizing interventions for their detection leads to generation of spurious modules in SCMs. We will argue that an SCM with a system of `only' \textit{a priori} equations is an analogue of the Bayesian Network (that is not necessarily acyclic). Therefore, a causal network containing \textit{a priori} and \textit{a posteriori} relations can be viewed as an intermediate between structural causal models and Bayesian networks.

The second inspection lies in section \ref{sec:invalidity}. In this section, we will examine two cases in which interventional inference of causation can violate the properties of the SCM in the natural situation. We will argue that the existence of these cases perplexes the generalization of interventionally inferred relationships to the natural domain. We will conclude our inspection by introducing the ``interventional'' and ``observational'' domains of causal networks and demonstrating a possible invalidity between these two contexts.

At the end of this paper and in section \ref{sec:final-remarks}, we will point out some remarks of this discussion. This section covers some supplementary discussion that are not fit in the main progression of this paper.

\section{Basic Notation}
\label{sec:basic-notation}
\subsection{Causal Terminology}
We make a distinction between relationship and relation. We use the term relationship more generally than relation -- a ``relationship'' between two phenomena shows that they are related, but the ``relation'' between them is the exact mathematical equation that governs them. For example, there is a relationship between the distance of two bodies and the gravitational force between them. However, the relation that defines how the gravitational force is related to the two bodies is Newton's law of universal gravitation.

We also use the terms causal detection and causal determination (identification) distinctively. The ``detection'' of a causal relationship between two phenomena is to prove that one is a cause of the other one. (This is a slightly different sense from that of the regularly-used terms ``causal discovery''. We use the term causal discovery to refer to the causal detection of each pair of phenomena in a system, i.e. the generation of the causal graph.) The ``determination'' of a causal relation is finding a mathematical equation that shows the exact effect of the causes of a phenomenon on it, namely the structural equation. We use the term ``inference'' to refer to both detection and determination collectively.

\subsection{Probabilistic Foundation}
We decompose probability spaces into a measurable space and its probability measure. We denote sets of outcomes using curly capital letters and their probability measures using the letter $\mathbbP$ subscripted with the corresponding set of outcomes. For example, $\mathbbP_\calX$ is a probability measure defined on the measurable space $\langle\calX , \calF_\calX\rangle$, with $\calF_\calX$ being a \textsigma-algebra defined on the set of outcomes $\calX$. We denote product spaces with curly capital boldface letters. As an example, $\langle\calbfX , \calbfF_\calbfX\rangle_\calI = \langle\prod_{i \in \calI}\calX_i, \bigotimes_{i \in \calI} \calF_{\calX_i}\rangle$ is the product space of $\langle\calX_i, \calF_{\calX_i}\rangle$ for $i \in \calI$, where $\calI$ is an index set. Also, in the general case, we use $\mathbbP\left(\cdot\right)$ as the function that measures the probability value of an event.

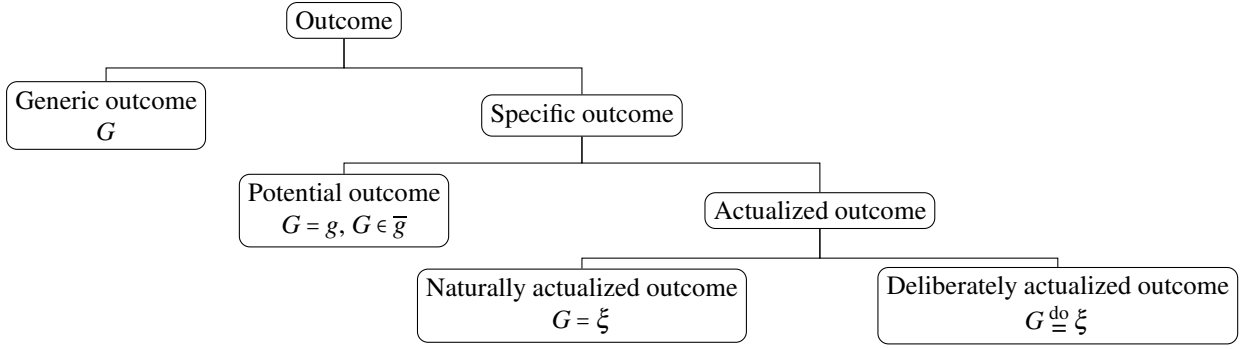
\begin{figure}
    \centering
    \begin{tikzpicture}[edge from parent fork down, level distance=8ex, sibling distance=18em, every node/.style = {shape=rectangle, rounded corners, draw, align=center}]
        \node {Outcome}
            child{node{Generic outcome\\$G$}}
            child{node{Specific outcome}
                child{node{Potential outcome\\$G = g$, $G \in \event{g}$}}
                child{node{Actualized outcome}
                    child{node{Naturally actualized outcome\\$G = \xi$}}
                    child{node{Deliberately actualized outcome\\$G \dooeq \xi$}}
                } 
            };
    \end{tikzpicture}
    \caption{Conceptual taxonomy of outcomes. The second line of each leaf node shows possible notation or operation for that outcome.}
    \label{fig:outcomes}
\end{figure}

Sometimes we consider an outcome without thinking of its possible values. We call such an outcome a \textit{generic outcome} and denote it using an italic capital letter. A generic outcome can attain possible values. We denote the possible value of an outcome with an italic small letter. When an outcome is considered along with its possible value, we call it a \textit{specific outcome}. However, as long as none of the possible values has taken place in the actual world, we are dealing with a \textit{potential outcome}. An \textit{actualized outcome} is a specific outcome whose possible value has taken place in the physical world. Indeed, when we talk about potential outcomes, we are considering an outcome along with its possible value, but the possible value itself is still indeterminate. But when we talk about actualized outcomes, that value is already fixed in the physical world. It should be noted that the potential/actual dichotomy is of physical importance and does not have any mathematical significance in this paper. Therefore, we will rarely distinguish between potential and actual outcomes and talk about them collectively as specific outcomes. Figure \ref{fig:outcomes} depicts the conceptual taxonomy of outcomes. As shown in the figure, an actualized outcome can be further divided into two new types. We deal with these types later in the paper. 

We use generic outcomes as an alternative to random variables. In one sense, generic outcomes are similar to random variables; they refer to the random happenings that are results of physical experiments. However, there are dissimilarities between them. Unlike random variables, generic outcomes are not measurable functions. A random variable can be thought of as a measurable function that maps the domain of one generic outcome to the domain of another generic outcome. Therefore, a generic outcome can be the output of a random variable. However, the domains of generic outcomes are not necessarily defined on $\mathbbR$. In this paper one can think of generic outcomes as random variables for an easier understanding. Curious readers can follow our defence on using this new terminology in the context of probabilistic causality in appendix \ref{sec:terminology}.

As an example for the taxonomy in the field of bioinformatics, $G_1$ and $G_2$ may be two generic outcomes representing the discretized expression of two genes,  with possible values $g \in \calX$ with $\calX = \left\{\text{low}, \text{medium}, \text{high}\right\}$. The equation $\mathbbP(G_1 = g) = \mathbbP(G_2 = g), \forall g \in \calX$ indicates that the expression of these two genes take the same value with the same probability for every possible value. The predicate $G_1 = \xi$ indicates that $G_1$ has attained a specific value in $\calX$. Also, please note that an analogues taxonomy can be made for events. However, the taxonomy of outcomes suffices in most cases. For example, an actualized event $\event{G_1} = \left\{\text{low}, \text{medium}\right\}$ can be simply represented using alternative notation $G_1 \in \left\{\text{low}, \text{medium}\right\}$. If necessary, we denote events using the same letters as outcomes but with a line above them.

We represent vectors of outcomes using boldface symbols. A vector is a set of outcomes indexed by a well-ordered index set. For example, $\mathbfit{G}_\calI = {\langle G_i \rangle}_{i \in \calI}$ is a vector of outcomes of expression values of genes $G_i$ ($i \in \calI$) that is ordered by the set $\calI$. We treat two vectors of outcomes as if they were plain sets and use them in conjunction with all set operands. 

When evident, we often drop the index set subscription of a product space or vector of outcomes and only use the boldface character, such as $\langle\calbfX, \calbfF_\calbfX\rangle$, or $\mathbfit{G}$.

\subsection{Graphical Foundation}
\label{graphical-representation}
Although causal models can be completely defined in the mathematical form, it is convenient to also have a graphical representation. The graphical representation of a causal model is typically a directed graph. We denote a graph by a pair $\calH = \langle \calV_\calH, \calD_\calH \rangle$ with $\calV_\calH$ being the set of nodes and $\calD_\calH$ being the set of directed edges. There is a directed edge from $i \in \calV_\calH$ to $j \in \calV_\calH$ if and only if $\left(i, j\right) \in \calD_\calH$. We denote this edge by $i \to j$ or $j \leftarrow i$.

We call $i$ a \textit{parent} of $j$ with respect to graph $\calH$ if and only if $\left(i, j\right) \in \calD_\calH$ and define the operation that returns all parents of $j$ as $\pa_\calH(j) \coloneqq \left\{i \in \calV_\calH \mid \left(i, j\right) \in \calD_\calH \right\}$. Similarly, node $i$ is a \textit{child} of node $j$ with respect to $\calH$ if and only if $\left(j, i\right) \in \calD_\calH$. We define the operation that returns all children of a node as $\ch_\calH(j) \coloneqq \left\{i \in \calV_\calH \mid \left(j, i\right) \in \calD_\calH \right\}$.

A directed path between two nodes $i_1$ and $i_n$ ($n > 1$) is a tuple $\langle i_1, \ldots, i_n \rangle$ such that $i_1, ..., i_n \in \calV_\calH$ and for each two adjacent elements $i_k$ and $i_{k+1}$ in the tuple, $i_k \in \pa_\calH(i_{k+1})$. We denote such a path by $i_1 \to \cdots \to i_n$. This leads to the definition of ancestors and descendants. We define the \textit{ancestors} of a node $j$ as $\an_\calH(j) \coloneqq \left\{i \in \calV_\calH \mid i = i_1 \to \cdots \to i_n = j \text{ and } \forall 1 \leq k < n : i_k \in \pa_\calH(i_{k+1}) \right\}$. Similarly, we define the \textit{descendants} of a node as $\de_\calH(j) \coloneqq \left\{i \in \calV_\calH \mid j = i_1 \to \cdots \to i_n = i \text{ and } \forall 1 \leq k < n : i_k \in \pa_\calH(i_{k+1}) \right\}$. The definition of ancestors and descendants can be generalized to encompass sets of nodes too. That is, $\an_\calH(\calI) \coloneqq \bigcup_{i \in \calI} \an_\calH(i)$ and $\de_\calH(\calI) \coloneqq \bigcup_{i \in \calI} \de_\calH(i)$. The same generalization is also applicable to the parents and children definitions.

\subsection{Structural Causal Models}
\label{mathematical-model}
In the most popular construction of SCM, the elements of the system are distinct generic outcomes that are divided into two types called exogenous (independent) and endogenous (dependent) outcomes \cite{bongers2016theoretical, mooij2016joint}. An exogenous generic outcome is not causally dependent on any other generic outcomes in the system. In contrast, the events of an endogenous generic outcome depend on the events of other exogenous or endogenous generic outcomes in the system.

Although there can be other formalizations, here, we follow the same mathematical models as proposed in \citen{mooij2016joint} with slight modifications. We represent an SCM as a tuple
$\calM = \langle \mathbfit{X}, \mathbfit{E}, \calH, \mathbff, \mathbbP_\calbfE \rangle$ comprised of:

\begin{enumerate}[label={(\roman*)}]
    \item a vector $\mathbfit{X} = \langle X_i \rangle_{i \in \calI}$ of endogenous generic outcomes belonging to a product measurable space $\langle \calbfX, \calbfF_\calbfX \rangle_\calI =  \langle\prod_{i \in \calI}\calX_i, \bigotimes_{i \in \calI} \calF_{\calX_i}\rangle$, 
    \item a vector $\mathbfit{E} = \langle E_j \rangle_{j \in \calJ}$ of exogenous generic outcomes belonging to a product measurable space $\langle \calbfE, \calbfF_\calbfE \rangle_\calJ =  \langle\prod_{j \in \calJ}\calE_j, \bigotimes_{j \in \calJ} \calF_{\calE_j}\rangle$, 
    \item a directed graph $\calH = \langle \calV_\calH, \calD_\calH \rangle$ representing the graphical causal relationships of the model, in which $\calV_\calH = \mathbfit{X} \cup \mathbfit{E}$ indicates the set of nodes and $\calD_\calH \subseteq \calV_\calH \times \mathbfit{X}$ indicates the set of edges,
    \item a measurable mapping $\mathbff : \calbfX \times \calbfE \to \calbfX$, such that each of its elements only depends on the corresponding subvector of outcomes determined by $\calH$:
    \begin{equation}
        \mathf_i : \calbfX_{\calI'} \times \calbfE_{\calJ'} \to \calX_i, \\
        \calI' = \left\{k \mid X_k \in \pa_\calH(X_i)\right\} \text{ and } \calJ' = \left\{k \mid E_k \in \pa_\calH(X_i)\right\}
        \text{,}
    \end{equation}
    \item and the joint probability measure $\mathbbP_\calbfE = \prod_{j \in \calJ} \mathbbP_{\calE_j}$ specifying the probability of exogenous events.
\end{enumerate}

We call $\calH$ the graph of $\calM$. In this graph every node represents a generic outcome. Nodes corresponding to the exogenous outcomes will not have any parents. Other nodes are descendants of the exogenous nodes. All nodes in $\an_\calH(X_i)$ are causes of $X_i$. The nodes in $\pa_\calH(X_i)$ are \textit{direct causes} of $X_i$.

We sometimes refer to $X_i = f_i(\mathbfit{X}, \mathbfit{E})$ as the \textit{structural equation} of node $X_i$ and use a system of structural equations to define the measurable mapping $\bm{f}$.

\section{Background}
\label{sec:background}
Two events $\event{X} \in \calF_\calX$ and $\event{Y} \in \calF_\calY$ from two measurable spaces $\langle\calX,\calF_\calX\rangle$ and $\langle\calY, \calF_\calY\rangle$ are thought as dependent if and only if $\mathbbP_{\calX\times\calY}\left({\event{X} \times \calY} \cap {\calX \times \event{Y}} \right) \neq \mathbbP_\calX\left(\event{X} \right) \mathbbP_\calY\left(\event{Y}\right)$. There is, however, no distinction between $\event{X}$ and $\event{Y}$ that would indicate that one must have been the cause of the other one. Three scenarios can be equally considered: 
\begin{enumerate*}[label={(\roman*)}] 
    \item that $\event{X}$ causes $\event{Y}$ (or equivalently, $\calY \setminus \event{Y}$),
    \item that $\event{X}$ (or $\calX \setminus \event{X}$) is caused by $\event{Y}$, and
    \item that both $\event{X}$ and $\event{Y}$ (or $\calY \setminus \event{Y}$) are caused by (at least) a third event (called a confounder).
\end{enumerate*}
These scenarios are not mutually exclusive, i.e. proving one does not automatically rule out either of the other two \cite{reichenbach1938experience,hofer1999reichenbach}.

In addition, by making a distinction between \textit{observational} (apparent) and \textit{actual} (underlying) probability measures of events, a fourth scenario comes into account, 
\begin{enumerate*}[resume,label={(\roman*)}] 
    \item where the actual measures of two random events are independent, but their observational measures indicate otherwise.
\end{enumerate*} We deem two reasons for the the discrepancy between the observational and actual measures:
\begin{enumerate*}[label={(\alph*)}] 
    \item the random or intentional sampling bias, which is known as internal invalidity, or
    \item the existence of extra (and probably hidden) \textit{extranous variables} that are falsely hypothesized not to be deviating from their actual distribution throughout the sampling procedure, known as the external invalidity.
\end{enumerate*}

We believe that this further distinction between internal and external invalidity is important, because one can be the result of erroneous sampling, but the other one can be attributed to the selection of a wrong SCM (e.g. not including a generic outcome in the model). As an example of external invalidity, in a two-dice experiment, if a hypothetical mechanism filters out the outcomes whose sum of values does not equal $7$, the statistician may identify a dependency between two variables, while there is no actual statistical dependence between the numbers appearing on the two dice.

By only observing the outcomes of random phenomena, it is often not determinant which of the four aforementioned scenarios are realized. In some cases, a lack of true knowledge about the actual causal relationships between phenomena can be problematic. A well-known example in the field of medical science statistics is Simpson's paradox \cite{wagner1982simpson,julious1994confounding,neuberg2003causality}. This paradox can result in performing the wrong treatment \cite{holt2016potential,franks2017post}.

A number of frameworks have been designed to overcome the problem of inferring causality from probabilistic dependence \cite{mellor1995facts,rosen1978defense,sekhon2008neyman}. Among them is the \textit{do-calculus}, which has arguably gained immense popularity. Firstly introduced by Judea Pearl, it is a mathematical system with the purpose to reconcile probability theory and causality. The basic idea of the do calculus is based on a dichotomy of \textit{observation} and \textit{intervention}. The do-calculus is therefore an axiomatic system for replacing post-interventional probability formulas with ordinary conditional probabilities. Pearl argues that intervention is a distinct process from observation and that this opens new doors to the problem of inferring causal relations, by means of experimentation.

In the next sub-section we will discuss the intuition behind the do-calculus and provide formal definition of interventions and their relation with causality inference.

\subsection{Interventions}
Pearl postulates the notion of \textit{intervention} and suggests that the primary way to discover a causal relationship between two generic outcomes is through interventions \cite{neuberg2003causality}. That is, we apply interventions to generic outcomes to discover a potential causal relationship between them. Although there are different types of interventions, we will keep focusing on the ``perfect interventions''. Henceforth, we use the terms intervetion and perfect intervention alternatively.

A \textit{perfect intervention} of an endogenous generic outcome is equivalent to enforcing it to attain a certain value, while preserving the probability measure of other outcomes. In the graphical representation, the perfect intervention of a node can be viewed as the removal of all of its parents and assignment of a specific value to it. Formally, a perfect intervention of a set of nodes $\mathbfit{X}_{\calI'} \subseteq \mathbfit{X}$ is defined as replacing the model $\calM = \langle \mathbfit{X}, \mathbfit{E}, \calH, \mathbff, \mathbbP_\calbfE \rangle$ with a modified model $\calM' = \langle \mathbfit{X}, \mathbfit{E}, \calH', \mathbff', \mathbbP_\calbfE \rangle$, where:
\begin{equation}
    X_i = \mathf'_i(\mathbfit{X}, \mathbfit{E}) = 
    \begin{cases}
        \xi_i & i \in \calI'\\
        \mathf_i(\mathbfit{X}, \mathbfit{E}) & \text{otherwise}
    \end{cases}
    \text{,}
\end{equation}
and the edges in $\calH'$ are pruned accordingly.

Pearl introduces the ``do'' operator as the probabilistic operator for perfect interventions. Here, we denote the do operator using the symbol ``$\dooeq$'' and use $X \dooeq \xi$ to represent the intervention where an outcome $X$ takes a specific value $\xi$. The distinction between $X = x$ and $X \dooeq \xi$ is depicted in figure \ref{fig:outcomes}. Both of these operators delineate the actualization of an outcome, but it is the perfect intervention that involves the deliberate enforcement of the outcome to take a specific value. 

Pearl's notion of interventions concludes that for an SCM $\calM$ with acyclic graph $\calH$, if there exists interventions ${X}_i \dooeq \xi$ and ${X}_i \dooeq \xi'$ with $\xi \neq \xi'$ such that $\mathbbP\left(X_j \in \event{x} \mid X_i \dooeq \xi \right) = \mathbbP\left(X_j \in \event{x} \mid X_i \dooeq \xi' \right)$ does not hold for events $\event{x} \in \calF_{\calX_j}$, then $X_i$ is a cause of $X_j$ with respect to $\calM$, i.e. $X_i \in \an_\calH(X_j)$ \cite{beebee2009oxford}.

Also, if $\calM$ is an SCM with acyclic $\calH$ and $\calI' = \calI \setminus \{j\}$, and if there exists interventions $\mathbfit{X}_{\calI'} \dooeq \bm{\xi}_{\calI'}$ and $\mathbfit{X}_{\calI'} \dooeq \bm{\xi}'_{\calI'}$ such that $\bm{\xi}_{\calI'\setminus\{i\}} = \bm{\xi}'_{\calI'\setminus\{i\}}$ and $\xi_{i} \neq \xi'_{i}$ such that $\mathbbP\left(X_j \in \event{x} \mid \mathbfit{X}_{\calI'} \dooeq \bm{\xi}_{\calI'} \right) = \mathbbP\left(X_j \in \event{x} \mid \mathbfit{X}_{\calI'} \dooeq \bm{\xi}'_{\calI'}  \right)$ does not hold for events $\event{x} \in \calF_{\calX_j}$, then $X_i$ is a direct cause of $X_j$ with respect to $\calM$, i.e. $X_i \in \pa_\calH(X_j)$ \cite{beebee2009oxford}. 

We call the two above rules Pearl's first and second rules of intervention, respectively. The rules of intervention also work for simple SCMs. A simple SCM is an SCM with a unique solution for every sub-system of structural equations for every endogenous and almost every exogenous potential outcome \cite{forre2017markov, mooij2018joint}. 

\section{{A Priori}--{A Posteriori} Distinction}
\label{sec:distinction}
We bring three examples that allow us to discuss the distinction: two hypothetical random experiments and a hypothetical causation model. The first random experiment is throwing a single dice and the second random experiment is counting each gender in a population. The hypothetical causation model belongs to the causal influence of water consumption in blood attenuation in humans.

\begin{figure}
    \centering
    \begin{tikzpicture}
        \begin{scope}[every node/.style={ellipse,draw,minimum height=3.5cm, minimum width=2cm}]
            \node (X) at (0.0, 0.0) {};
            \node (Y) at (5.0, 0.0) {};
            \node (Z) at (10, 0.0) {};
        \end{scope}
        \begin{scope}[yshift=-1.8ex]
            \node at (0.0, -1.75) {$\calY$};
            \node at (5.0, -1.75) {$\calX$};
            \node at (10.0, -1.75) {$\calZ$};
        \end{scope}
        \begin{scope}[xshift=5cm, every node/.style={ellipse,draw,fill=black,inner sep=0.05cm}, label distance=-1mm]
            \node [label=$1$] (N1) at (-0.1, 1.1) {};
            \node [label=$2$] (N2) at (+0.1, 0.6) {};
            \node [label=$3$] (N3) at (-0.1, 0.1) {};
            \node [label=$4$] (N4) at (+0.1, -0.4) {};
            \node [label=$5$] (N5) at (-0.1, -0.9) {};
            \node [label=$6$] (N6) at (+0.1, -1.4) {};
        \end{scope}
        \begin{scope}[every node/.style={ellipse,draw,fill=black,inner sep=0.05cm}, label distance=1mm]
            \node [label=odd] (odd) at (0.0, 0.8) {};
            \node [label=even] (even) at (0.0, -0.8) {};
        \end{scope}
        \begin{scope}[xshift=10cm, every node/.style={ellipse,draw,fill=black,inner sep=0.05cm}, label distance=1mm]
            \node [label=$\leq3$] (Leq3) at (0.0, 0.8) {};
            \node [label=$>3$] (G3) at (0.0, -0.8) {};
        \end{scope}
        \begin{scope}[every edge/.style={-, draw}]
            \path [bend left=10] (odd) edge node {} (N1);
            \path [bend left=10] (odd) edge node {} (N3);
            \path [bend left=10] (odd) edge node {} (N5);
            \path [bend right=10] (even) edge node {} (N2);
            \path [bend right=10] (even) edge node {} (N4);
            \path [bend right=10] (even) edge node {} (N6);
        \end{scope}
        \begin{scope}[every edge/.style={-, draw}]
            \path [bend right=10] (Leq3) edge node {} (N1);
            \path [bend right=10] (Leq3) edge node {} (N2);
            \path [bend right=10] (Leq3) edge node {} (N3);
            \path [bend left=10] (G3) edge node {} (N4);
            \path [bend left=10] (G3) edge node {} (N5);
            \path [bend left=10] (G3) edge node {} (N6);
        \end{scope}
    \end{tikzpicture}
    \caption{The mapping between three sets of outcomes for the experiment of throwing a dice.}
    \label{fig:dice-outcomes}
\end{figure}
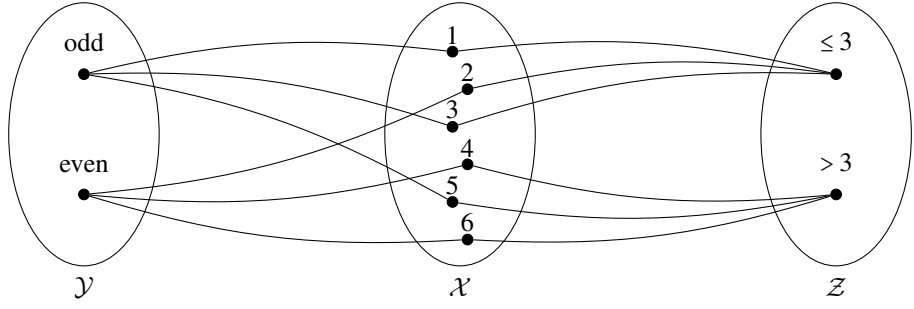

It is possible consider many outcome spaces for the dice-throwing experiment. Three of them are shown in figure \ref{fig:dice-outcomes}. The sets of outcomes are officially defined as $\calX = \{1, 2, 3, 4, 5, 6\}$, $\calY = \{\text{odd}, \text{even}\}$, $\calZ = \{{\leq3}, {>3}\}$. We also let $X$, $Y$ and $Z$ be three generic outcomes defined on these three domains.

Similarly, many outcome spaces are considerable for counting each gender inside a population. We consider three outcome spaces with three corresponding generic outcomes: number of females $G$ defined on the set of outcomes $\calG = \{0, 1, 2, ...\}$, number of males $B$ defined on the set of outcomes $\calB = \{0, 1, 2, ...\}$, and the size of population $P$ defined on $\calP = \{0, 1, 2, ...\}$. Since in the real world there might be other genders in the population, it is expected that the size of population is \textit{approximately} equal to the number of females plus the number of males. Therefore, we also define the number of non-women $N$ on $\calN = \{0, 1, 2, ...\}$.

We consider two generic outcomes for the hypothetical causation model: The amount of water drank $\calW$ defined on the space $\langle \mathbbR, \calB(\mathbbR) \rangle$ and the blood attenuation $\calA$ defined on $\langle \mathbbR, \calB(\mathbbR) \rangle$, respectively in milliliters and Hounsfield unit. The structural equations of the SCM representing the model are:
\begin{equation}
    \begin{cases}
        W = f_1(E_1) = E_1\\
        A = f_2(W, E_2) = \alpha W + E_2
    \end{cases}\text{,}
\end{equation}
where $E_1$ and $E_2$ are the exogenous generic outcomes and $\alpha$ is a constant coefficient.

\subsection{A Priori Relations}
\textit{A priori} (`from the earlier') and \textit{a posteriori} (`from the later') are two Greek terms popularized by Immanuel Kant to make a distinction between propositions that are known prior to experience and those that are not. Although the truth and domain of such a distinction have been matters of debate \cite{machery2006two}, we take these two concept for granted, and extend them to SCMs. Formally, \textit{a priori} knowledge or justification is independent of experience, as with mathematical relations (3 + 2 = 5) or tautologies (``All bachelors are unmarried''). In contrast, \textit{a posteriori} knowledge or justification depends on experience or empirical evidence, as with most aspects of science and personal knowledge \cite{smith2003commentary, lewis1923pragmatic}.

We maintain these definitions as proposed, and apply them to the structural equations in the SCM, i.e. the measurable mappings defined for generic outcomes, or the functions $\mathit{f}_i$. In general, in an SCM $\calM = \langle \mathbfit{X}, \mathbfit{E}, \calH, \mathbff, \mathbbP_\calbfE \rangle$, we call a measurable mapping $\mathf_i : \calbfX_{\calI'} \times \calbfE_{\calJ'} \to \calX_i$ \textit{a priori} if and only if:
\begin{enumerate}[label={(\roman*)}]
    \item for every vector of potential outcomes $\langle \bm{x}_{\calI'}, \bm{e}_{\calJ'} \rangle$, $\mathf_i$ returns the potential outcome $x_i$ \textit{a priori},
    \item and if $\calJ' \neq \emptyset$, then $\mathbbP_{\calbfE_{\calJ'}}$ is \textit{a priori} known.
\end{enumerate}
Based on criterion (i), measurable functions like $f_i(\bm{x}, \bm{e}) = g\left(h_1(\bm{x}, \bm{e}), ..., h_n(\bm{x}, \bm{e})\right)$ are not \textit{a priori} if at least one function $h_i$ ($i \in \left[1, n\right]$) is not \textit{a priori} known. Without criterion (ii) any irrelevant function may be thought of as \textit{a priori}. As an exemplary equation, consider $X = f(Y, E) = Y + E$, without a known measure $\mathbbP_\calE$. $f$ can be true regardless of what $X$ and $Y$ refer to, because for every $X$ and $Y$ there is one probability measure such that $X = f(Y, E)$ holds. This means that without a distinguishing probability measure of the incoming exogenous generic outcomes, $f$ bears no real knowledge.

In our population example, the size of the population is \textit{a priori} known to be the sum of females and non-females. That is, $P = f_1(G, N) = G + N$ is an \textit{a priori} structural equation. On the contrary, $P = f_2(G, N, E) = G + B + E$ is \textit{a posteriori} known, because the true relation between population size ($P$) and number of females and males ($G$ and $B$) is only determined after figuring out the true measure of error ($E$).

\begin{figure}
    \centering
    \subcaptionbox{\label{fig:blood-experiment-a}}{
        \begin{tikzpicture}
            \node (c1) at (-4.0, 0.0) {};
            \node (c2) at (+4.0, 0.0) {};
            \begin{scope}[every node/.style={circle,thick,draw,minimum size=0.8cm}]
                \node (A) at (0.0, -1.0) {$A$};
                \node (A_1) at (-2.0, 1.0) {$A_1$};
                \node (A_2) at (2.0, 1.0) {$A_2$};
                \node (W_1) at (-1.0, 2.0) {$W_1$};
                \node (W_2) at (1.0, 2.0) {$W_2$};
            \end{scope}
            \begin{scope}[every node/.style={circle,draw,inner sep=0,minimum size=0.5cm}]
                \node (E_1) at (-2.0, 3.0) {\small $E_1$};
                \node (E_2) at (+2.0, 3.0) {\small $E_2$};
                \node (E_3) at (-3.0, 2.0) {\small $E_3$};
                \node (E_4) at (+3.0, 2.0) {\small $E_4$};
            \end{scope}
            \begin{scope}[every edge/.style={->, thick, draw}]
                \path (E_1) edge node {} (W_1);
                \path (E_2) edge node {} (W_2);
                \path (E_3) edge node {} (A_1);
                \path (E_4) edge node {} (A_2);
                \path (W_1) edge node {} (A_1);
                \path (W_2) edge node {} (A_2);
                \path (A_1) edge node {} (A);
                \path (A_2) edge node {} (A);
            \end{scope}
        \end{tikzpicture}
    }
    \subcaptionbox{\label{fig:blood-experiment-b}}{
        \begin{tikzpicture}
            \node (c1) at (-4.0, 0.0) {};
            \node (c2) at (+4.0, 0.0) {};
            \begin{scope}[every node/.style={circle,thick,draw,minimum size=0.8cm}]
                \node (A) at (0.0, -1.0) {$A$};
                \node (A_1) at (-2.0, 1.0) {$A_1$};
                \node (W_2) at (2.0, 1.0) {$W_2$};
                \node (W_1) at (-1.0, 2.0) {$W_1$};
            \end{scope}
            \begin{scope}[every node/.style={circle,draw,inner sep=0,minimum size=0.5cm}]
                \node (E_1) at (-2.0, 3.0) {\small $E_1$};
                \node (E_2) at (+3.0, 2.0) {\small $E_2$};
                \node (E_3) at (-3.0, 2.0) {\small $E_3$};
                \node (E_4) at (0.0, 1.0) {\small $E_4$};
            \end{scope}
            \begin{scope}[every edge/.style={->, thick, draw}]
                \path (E_1) edge node {} (W_1);
                \path (E_2) edge node {} (W_2);
                \path (E_3) edge node {} (A_1);
                \path (E_4) edge node {} (A);
                \path (W_1) edge node {} (A_1);
                \path (W_2) edge node {} (A);
                \path (A_1) edge node {} (A);
            \end{scope}
        \end{tikzpicture}
    }
    
    \caption{Hypothetical causal model for averaging blood attenuation of two persons. $W_i$ and $A_i$ are respectively the water consumption and blood attenuation of person $i$. $A$ is the average blood attenuation. \subref{fig:blood-experiment-a} The un-marginalized network. \subref{fig:blood-experiment-b} The same network after marginalizing $A_2$.}
    \label{fig:blood-experiment}
\end{figure}
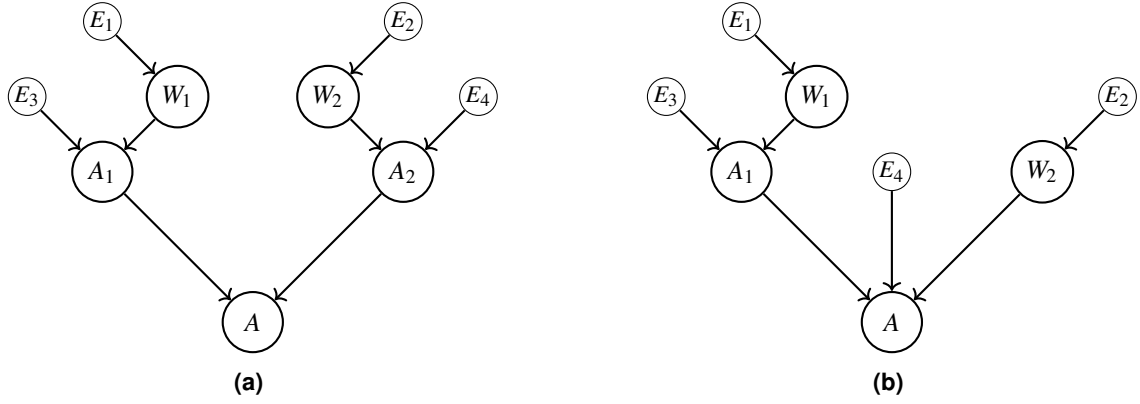

\subsection{Marginalization of A Priori Relations}
\label{sec:marginalization}
SCMs have the property of (de-)marginalization \cite{bongers2016theoretical}. This property allows us to change the structural equations of SCMs. (De-)marginalization may also change the \textit{a priority}/\textit{a posteriority} property of altered strutural equations. Consider figure \ref{fig:blood-experiment} \subref{fig:blood-experiment-a} as an example based on our hypothetical causation model, where a generic outcome $A$ is the average blood attenuation of two people. The system of structural equations of this SCM is as follows:
\begin{equation}
    \begin{cases}
        W_1 = f_1(E_1) = E_1\\
        W_2 = f_2(E_2) = E_2\\
        A_1 = f_3(W_1, E_3) = \alpha_1  W_1 + E_3\\
        A_2 = f_4(W_2, E_4) = \alpha_2 W_2 + E_4\\
        A = f_5(A_1, A_2) = \frac{1}{2} \times (A_1 + A_2)\\
    \end{cases}
    \text{,}
\end{equation}
with $\alpha_1$ and $\alpha_2$ being two constants. In this system, $f_1$ and $f_2$ are \textit{a posteriori} because they contradict criterion (ii), $f_3$ and $f_4$  are \textit{a posteriori} because they contradict criteria (i) and (ii), and $f_5$ is \textit{a priori}. By marginalizing $A_2$, we can make a new smaller SCM as shown in figure \ref{fig:blood-experiment} \subref{fig:blood-experiment-b} with the following system of structural equations:
\begin{equation}
    \begin{cases}
        W_1 = f'_1(E_1) = E_1\\
        W_2 = f'_2(E_2) = E_2\\
        A_1 = f'_3(W_1, E_3) = \alpha_1 W_1 + E_3\\
        A = f'_5(A_1, W_2, E_4) = \frac{1}{2} \times (A_1 + \alpha_2 W_2 + E_4)\\
    \end{cases}
\end{equation}
The relationship defining $A$ used to be \textit{a priori} in the un-marginalized SCM. However, the new relationship, that is $f'_5$, is known only after experimentation; it is not an \textit{a priori} relationship anymore. The meaning of the generic outcome $A$ is ``the average amount of blood attenuation of the first person and water consumption of the second person, which is the linear cause of blood attenuation of the second person''. This is a different sense from that of the previous SCM and is not determined before experimentation.

\subsection{Interventional Detection of A Priori Relations}
Pearl's rules of intervention are able to indicate causal \textit{a periori} relationships. However, \textit{a priori} relations are antithetical to causal relations as known to human-being. The reason is twofold. On the one hand, a causal relationship is detected \textit{a posteriori}. Ordinarily, an experimenter refers to at least two physical phenomena in the external world, measures two generic outcomes belonging to those two phenomena, and makes causal inferences based on those generic outcomes. For example, the relationship between water consumption and blood attenuation requires the observer to collect two statistical data: one being the water consumption values and the other one being the corresponding blood attenuation values.

On the other hand, causal relationships are asymmetric in essence. [Excluding temporal systems], for two phenomena $A$ and $B$, either $A$ is the cause of $B$, or $B$ is the cause of $A$. However, structural equations do not have any inherit asymmetry. For example, in the experiment of counting the number of each gender in the population, there is no causal distinction between $P = G + N$ (population size is caused by number of females and number of non-females) and $G = P - N$ (number of females is caused by population size and number of non-females), even though such a equation may have ``constitutional asymmetry'' \cite{zangwill2012constitution}. This is also true for temporal causal models as their cycles can be unfolded in time.

It is noteworthy that although \textit{a priori} relations are symmetric in terms of causality, Pearl's second rule of intervention does not necessarily discover \textit{a priori} edges symmetrically. That is, if $f_i$ is \textit{a priori} and $X_j \in \pa_\calH(X_i)$, then 
\begin{equation}
    (X_j, X_i) \in_\text{intervention} \calD_\calH \Leftrightarrow (X_i, X_j) \in_\text{intervention} \calD_\calH
    \label{eq:bidirectional-edge}
\end{equation}
does not hold. Here, we use $\in_\text{intervention}$ to denote an edge that is detectable by applying Pearl's second rule of intervention.

Let us consider the experiment of throwing a dice. If we want to form a causal network consisting of only $Y$ and $Z$, the system of structural equations can be as follows:
\begin{equation}
    \begin{cases}
        Z = f_1(Y, E_1) = \begin{cases}
            {\leq 3} & Y = \text{odd} \text{ and } E_1 = 0\\
            {> 3} & Y = \text{odd} \text{ and } E_1 = 1\\
            {\leq 3} & Y = \text{even} \text{ and } E_1 = 1\\
            {> 3} & Y = \text{even} \text{ and } E_1 = 0\\
        \end{cases}\\
        Y = f_2(Z, E_2) = \begin{cases}
            \text{odd} & Z \leq 3 \text{ and } E_2 = 0\\
            \text{even} & Z > 3 \text{ and } E_2 = 1\\
            \text{odd} & Z \leq 3 \text{ and } E_2 = 1\\
            \text{even} & Z > 3 \text{ and } E_2 = 0\\
        \end{cases}\\
    \end{cases}\\
    \text{,}\\
    E_1, E_2 \sim \text{Ber}(\frac{1}{3})
\end{equation}
Both $f_1$ and $f_2$ are \textit{a priori}. Pearl's second rule of causation may or may not infer a causal edge from $Y$ to $Z$ and vice versa. If $\mathbbP\left(Z \leq 3 \mid Y \dooeq \text{odd} \right) = \frac{2}{3}$ and $\mathbbP\left(Z \leq 3 \mid Y \dooeq \text{even} \right) = \frac{1}{3}$, then $Y$ is detected as a cause of $Z$. However, if $Y \dooeq \text{odd}$ and  $Y \dooeq \text{even}$ are respectively obtained by enforcing the dice to take either of the values in $\{1, 5\}$ and $\{2, 6\}$, then both of the previous measures will be equal to $\frac{1}{2}$ and no causal edge from $Y$ to $Z$ will be inferred. It is possible that both, only one, or neither of the edges $(Y, Z)$ and $(Z, Y)$ are detected.

We call an intervention $X_i \dooeq \xi_i$ \textit{consistent} with respect to $X_j$, if and only if $\mathbbP\left(X_j \in \event{x_j} \mid X_i \dooeq \xi_i \right) = \mathbbP\left(X_j \in \event{x_j} \mid X_i = \xi_i \right)$. If $\calM = \langle \mathbfit{X}, \mathbfit{E}, \calH, \mathbff, \mathbbP_\calbfE \rangle$ is a simple SCM on which intervention $\mathbfit{X}_{\calI\setminus\{i, j\}} \dooeq \bm{\xi}_{\calI\setminus\{i, j\}}$ is applied, then if $X_i \dooeq \xi_i$ and $X_j \dooeq \xi_j$ are consistent for almost every $\xi_i \in \calX_i$ and $\xi_j \in \calX_j$, then the statement in equation \ref{eq:bidirectional-edge} holds. To prove this, let us assume that $\calI' = \calI\setminus\{i, j\}$. Then, it is sufficient to note that:
\begin{equation}
    \begin{split}
        X_j \not\bigCI X_i \mid \mathbfit{X}_{\calI'} \dooeq \bm{\xi}_{\calI'} \Leftrightarrow
        \exists \xi_i', \xi_i'' : 
        \mathbbP\left(X_j \in \event{x_j} \bigm\mid X_i = \xi'_i, \mathbfit{X}_{\calI'} \dooeq \bm{\xi}_{\calI'} \right) \neq \mathbbP\left(X_j \in \event{x_j} \bigm\mid X_i = \xi''_i, \mathbfit{X}_{\calI'} \dooeq \bm{\xi}_{\calI'} \right) \\\Leftrightarrow
        \exists \xi_i', \xi_i'' : 
        \mathbbP\left(X_i \in \event{x_i} \bigm\mid X_j = \xi'_j, \mathbfit{X}_{\calI'} \dooeq \bm{\xi}_{\calI'} \right) \neq \mathbbP\left(X_i \in \event{x_i} \bigm\mid X_j = \xi''_j, \mathbfit{X}_{\calI'} \dooeq \bm{\xi}_{\calI'} \right)
    \end{split}
\end{equation}
Since $X_i \dooeq \xi_i$ and $X_i \dooeq \xi_i$ are consistent under the condition that $\mathbfit{X}_{\calI\setminus\{i, j\}} \dooeq \bm{\xi}_{\calI\setminus\{i, j\}}$, then each of the inequality statements of the above equation provides the sufficient condition for one direction of the implication in equation \ref{eq:bidirectional-edge}.

Since the structural equation of an \textit{a priori} node is symmetric, it is tempting to discuss whether the links to an \textit{a priori} node are reversible or not. Although these relations are not causal, the incoming edges to such a node are not always reversible. If $f_i$ is \textit{a priori}, it is expected that none of its parents are real causes of $X_i$, because all information that $f_i$ holds is known before experimentation. If $X_i$ is a node with \textit{a priori} relationship in $\calH$, then an ``\textit{a priori} module'' is formed around it. By carefully adding or removing the edges that lie inside or cross this module, the module can be re-configured in a way that the resulting SCM is equivalent to the original SCM. In appendix \ref{sec:reversibility} we will discuss the possible configuration of \textit{a priori} modules.

\subsection{Faithfulness Property of SCMs with A Priori Relations}
Let $\calM = \langle \mathbfit{X}, \mathbfit{E}, \calH, \mathbff, \mathbbP_\calbfE \rangle$ be an acyclic SCM. $\calM$ is faithful if and only if for all three distinct subsets $\mathbfit{X}_1$, $\mathbfit{X}_2$, $\mathbfit{X}_3$ of $\mathbfit{X}$:
\begin{equation}
    \mathbfit{X}_1 \bigCI_{\mathbbP_\calbfE} \mathbfit{X}_2 \mid \mathbfit{X}_3 \\\Rightarrow\\ \mathbfit{X}_1 \bot_{\calH} \mathbfit{X}_2 \mid \mathbfit{X}_3 \text{ ,}
\end{equation}
where $\mathbfit{X}_1 \bigCI_{\mathbbP_\calbfE} \mathbfit{X}_2 \mid \mathbfit{X}_3$ shows the probabilistic independence of $\mathbfit{X}_1$ and $\mathbfit{X}_2$ conditioned on $\mathbfit{X}_3$, and $\mathbfit{X}_1 \bot_{\calH} \mathbfit{X}_2 \mid \mathbfit{X}_3$ shows that $\mathbfit{X}_1$ and $\mathbfit{X}_2$ are d-separated by $\mathbfit{X}_3$.

Although acyclic SCM do not always hold the property of faithfulness, they typically maintain it to the measure zero sets of parameters \cite{meek2013strong}. This means that in order to violate the faithfulness property of an SCM, the parameters of the structural relations have to be tuned ultimately thoroughly. Therefore, the faithfulness property is almost surely preserved in networks with only \textit{a posteriori} relations. But for \textit{a priori} relations, it is easy to form fine-tuned relations and therefore this property is easily violated.

\section{Cross-Domain Invalidity}
\label{sec:invalidity}
As pointed out in section \ref{sec:background}, a major problem with causal detection using probabilistic dependence is that this method can not distinguish between straight causation and confounding effects. Even if an observer is justified about the true causal relationship between physical phenomena (i.e. he/she can choose between scenarios (i) through (iii)), and he/she is assured that there is no sampling bias (scenario (iv-a)), it is still perverse to believe that the resulted model is externally valid (scenario (iv-b)).

Interventional detection of causal relationships is not immune to similar errors either. In this section, we will show that the interventional detection of causal relationships is based on hidden assumptions that are not universally true. Henceforth, we will also explicitly distinguish exogenous generic outcomes, that are the source of randomness in the model of the system, and extraneous generic outcomes, that are not part of the constructed model of the system.

Similar to Pearl's original nomenclature, we will use the notions of front-door and back-door effects. Based on this dichotomy, we classify the problems that may arise when using interventions to discover causation into two classes. We call these two classes the front-door interventional interference and back-door interventional interference. Each of these problems may cause the network that is inferred under intervention not to be valid in the natural situation. This is important, because there are cases that an experimenter is doing interventions to infer the natural causal relationships. We consider two domains for a network: interventional domain, which is the context where the network is under intervention, and observational domain, where the network is not under intervention. A ``domain'' is nothing more than a context; a domain is a set of conditions in which the network is observed. We use the term domain in order to avoid confusion with the ordinary usage of contexts.

We define two mutually exclusive domains for a causal system: 
\begin{enumerate*}[label={(\roman*)}]
    \item the domain where a causal system is observed under interventions, which we call the \textit{interventional domain} of the network, and
    \item the domain where the network is observed without application of any interventions, which we call the \textit{natural domain} of the network.
\end{enumerate*}
We call the consistency of the inferred network under interventions with the natural network the \textit{cross-domain validity}. The experimenter should be cautious about the possible inconsistency if what he/she is going to infer via rules of intervention is the natural behaviour of the system. 

\begin{figure}
    \centering
    \subcaptionbox{\label{fig:cross-domain-validity-a}}{
        \begin{tikzpicture}
            \node at (-1.0, 0.0) {};
            \node at (5.0, 0.0) {};
            \node (int) at (0, 1.0) {intervention};
            \begin{scope}[every node/.style={circle,thick,draw,minimum size=0.8cm}]
                \node (X_1) at (0.0, 0.0) {$X_i$};
                \node (X_2) at (2.0, 0.8) {...};
                \node (X_3) at (4.0, 0.0) {$X_j$};
            \end{scope}
            \begin{scope}[every node/.style={circle, dashed,thick,draw,minimum size=0.8cm}]
                \node (X_m) at (0, -1.5) {$H_m$};
            \end{scope}
            \begin{scope}[every edge/.style={->, thick, draw}]
                \path [bend right=10] (X_1) edge node {} (X_2);
                \path [bend left=10] (X_2) edge node {} (X_3);
                \path [dashed] (X_1) edge node {} (X_m);
                \path [bend right=30, dashed] (X_m) edge node {} (X_3);
                \path (int) edge node {} (X_1);
            \end{scope}
        \end{tikzpicture}
    }
    \subcaptionbox{\label{fig:cross-domain-validity-b}}{
        \begin{tikzpicture}
            \node at (-1.0, 0.0) {};
            \node at (5.0, 0.0) {};
            \node (int) at (0, 2.5) {intervention};
            \begin{scope}[every node/.style={circle,thick,draw,minimum size=0.8cm}]
                \node (X_1) at (0.0, 0.0) {$X_i$};
                \node (X_2) at (2.0, -0.8) {...};
                \node (X_3) at (4.0, 0.0) {$X_j$};
            \end{scope}
            \begin{scope}[every node/.style={circle, dashed,thick,draw,minimum size=0.8cm}]
                \node (X_m) at (0, 1.5) {$H_m$};
            \end{scope}
            \begin{scope}[every edge/.style={->, thick, draw}]
                \path [bend left=10] (X_1) edge node {} (X_2);
                \path [bend right=10] (X_2) edge node {} (X_3);
                \path [dashed] (X_m) edge node {} (X_1);
                \path [bend left=30, dashed] (X_m) edge node {} (X_3);
                \path (int) edge node {} (X_m);
            \end{scope}
        \end{tikzpicture}
    }
    
    \caption{Two types of interventional interference. \subref{fig:cross-domain-validity-a} Front-door interventional interference. \subref{fig:cross-domain-validity-b} Back-door interventional interference. In each of these sub-figures, an intervention manipulates a door and alters it to an unnatural state. The door is specified by an extraneous generic outcome labeled $H_m$. This leads to an invalid detection of causation between $X_i$ and $X_j$. Dashed nodes are extraneous generic outcomes. Nodes indicated with ``$\cdots$'' show a compact schematic of part of the network.}
    \label{fig:cross-domain-validity}
\end{figure}
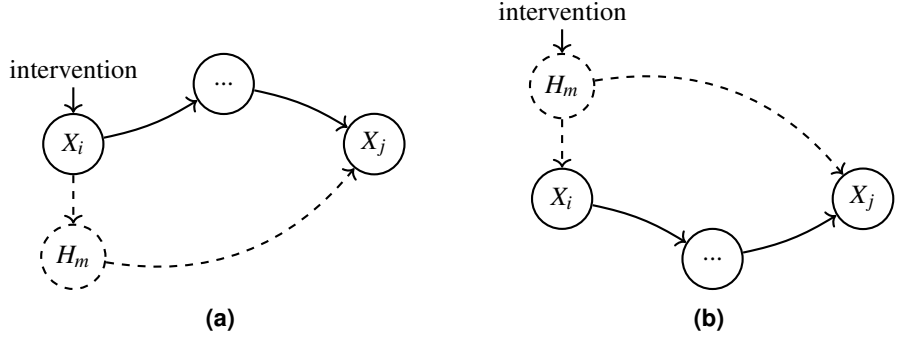

Each of the aforementioned classes of problems causes a cross-domain invalidity problem, i.e. makes the network that is inferred in the interventional domain invalid in the natural situation. Based on which class has caused the invalidity, we will divide the cross-domain invalidity into two types: 
\begin{enumerate}[label={(\roman*)}]
    \item Front-door cross-domain invalidity, caused by front-door interventional interference, where the intervention of $X_i$ alters a front door to $X_j$ to a state different from that of the natural situation. An exemplary generic schema of this interference is depicted in as in figure \ref{fig:cross-domain-validity} \subref{fig:cross-domain-validity-a}.
    \item Back-door cross-domain invalidity, caused by back-door interventional interference, where the intervention of $X_i$ alters a back door to $X_j$ to a state different from that of the natural situation. An exemplary generic schema of this interference is depicted in as in figure \ref{fig:cross-domain-validity} \subref{fig:cross-domain-validity-b}.
\end{enumerate}
In the following sub-sections, we will explain how each of these interference effects are realized. 

\subsection{Front-Door Cross-Domain Invalidity}
The front-door cross-domain invalidity takes place if the interventional detection of the causal relationship between $X_i$ and $X_j$ is accompanied by a front-door interventional interference. If an intervention of $X_i$ undermines the value of $X_j$ through an undiscovered front-door from $X_i$ to $X_j$, then we say that a front-door interventional interference has taken place.

Imagine an experimenter conducting an experiment in in the field of bioinformatics. He/She wants to examine the causal relationship between two generic outcomes in a gene regulatory network. Figure \ref{fig:grn-causation} \subref{fig:grn-causation-a} depicts the system under this examination. This system is comprised of four genes ($G_1, ..., G_4$) and three transcription factors ($T_2, T_3, T_4$). In this SCM, each of the generic outcomes show the concentration level of the mRNA or protein related to the corresponding element of the system. Note that the system has two extraneous generic outcomes $G_4$ and $T_4$. The corresponding model to this hypothetical system has the following system of structural equations:
\begin{equation}
    \begin{cases}
        G_1 = f_1(E_1) = E_0 + E_1\\
        T_2 = f_2(G_1, E_2) = G_1 + E_2\\
        G_2 = f_3(T_2, E_2) = 1 - T_2 + E_3\\
        T_3 = f_4(G_2, E_4) = G_2 + E_4\\
        G_3 = f_5(T_3, E_5) = T_3 + E_5\\
    \end{cases}
    \text{,}
\end{equation}
with $E_0 \sim \text{Ber}(\frac{1}{2})$ and $E_i \sim \text{Normal}(0, \sigma_i)$ ($i = 1, ..., 7$). Suppose that the causal relationship between the generic outcomes of this model is based on the chemical study of the involved compounds. Also, the structural equations are estimates based on how these elements work chemically. $T_2$ acts as a suppressor of $G_2$ and $T_3$ acts as an activator of $G_3$. Besides these elements, there is the ``undiscovered'' gene $G_4$ that is only activated when it is exposed to high doses of transcription factors $T_2$ and $T_3$. This gene is never expressed naturally and its value is always almost zero. But if it is activated, its production suppresses $G_3$. This means that the actual system has the following system of structural equations:
\begin{equation}
    \begin{cases}
        G_1 = f_1(E_1) = E_0 + E_1\\
        T_2 = f_2(G_1, E_2) = G_1 + E_2\\
        G_2 = f_3(T_2, E_2) = 1 - T_2 + E_3\\
        T_3 = f_4(G_2, E_4) = G_2 + E_4\\
        G_3 = f_5'(T_3, E_5) = (T_3 \lor T_4) + E_5\\
        G_4 = f_6(T_2, T_3, E_6) = (T_2 \land T_3) + E_6\\
        T_4 = f_7(G_4, E_7) = G_4 + E_7
    \end{cases}
\end{equation}
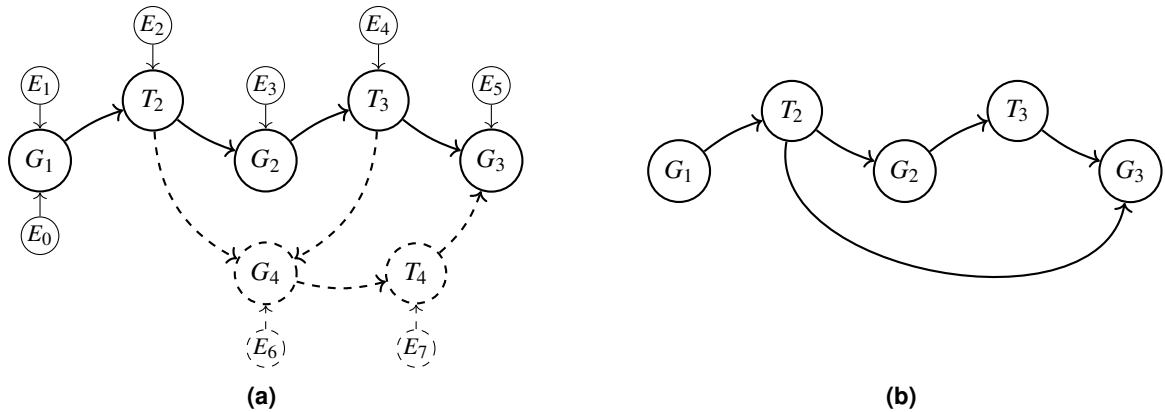
\begin{figure}
    \centering
    \subcaptionbox{\label{fig:grn-causation-a}}{
        \begin{tikzpicture}
            \node at (-1.0, 0.0) {};
            \node at (7.0, 0.0) {};
            \begin{scope}[every node/.style={circle,thick,draw,minimum size=0.8cm}]
                \node (G_1) at (0.0, 0.0) {$G_1$};
                \node (T_2) at (1.5, 0.8) {$T_2$};
                \node (G_2) at (3.0, 0.0) {$G_2$};
                \node (T_3) at (4.5, 0.8) {$T_3$};
                \node (G_3) at (6.0, 0.0) {$G_3$};
            \end{scope}
            \begin{scope}[every node/.style={circle, dashed,thick,draw,minimum size=0.8cm}]
                \node (G_4) at (3.0, -1.5) {$G_4$};
                \node (T_4) at (5.0, -1.5) {$T_4$};
            \end{scope}
            \begin{scope}[every node/.style={circle,draw,inner sep=0,minimum size=0.5cm}]
                \node (E_0) at (0.0, -1.0) {\small $E_0$};
                \node (E_1) at (0.0, 1.0) {\small $E_1$};
                \node (E_2) at (1.5, 1.8) {\small $E_2$};
                \node (E_3) at (3.0, 1.0) {\small $E_3$};
                \node (E_4) at (4.5, 1.8) {\small $E_4$};
                \node (E_5) at (6.0, 1.0) {\small $E_5$};
                \node [dashed] (E_6) at (3.0, -2.5) {\small $E_6$};
                \node [dashed] (E_7) at (5.0, -2.5) {\small $E_7$};
            \end{scope}
            \begin{scope}[every edge/.style={->, thick, draw}]
                \path [bend left=10] (G_1) edge node {} (T_2);
                \path [bend right=10] (T_2) edge node {} (G_2);
                \path [bend left=10] (G_2) edge node {} (T_3);
                \path [bend right=10] (T_3) edge node {} (G_3);
                \path [bend right=30, dashed] (T_2) edge node {} (G_4);
                \path [bend left=30, dashed] (T_3) edge node {} (G_4);
                \path [bend right=15, dashed] (G_4) edge node {} (T_4);
                \path [bend right=15, dashed] (T_4) edge node {} (G_3);
            \end{scope}
            \begin{scope}[every edge/.style={->, draw}]
                \path (E_0) edge node {} (G_1);
                \path (E_1) edge node {} (G_1);
                \path (E_2) edge node {} (T_2);
                \path (E_3) edge node {} (G_2);
                \path (E_4) edge node {} (T_3);
                \path (E_5) edge node {} (G_3);
                \path [dashed] (E_6) edge node {} (G_4);
                \path [dashed] (E_7) edge node {} (T_4);
            \end{scope}
        \end{tikzpicture}
    }
    \subcaptionbox{\label{fig:grn-causation-b}}{
        \begin{tikzpicture}
            \node at (-1.0, 0.0) {};
            \node at (7.0, 0.0) {};
            \node at (0.0, -2.5) {};
            \begin{scope}[every node/.style={circle,thick,draw,minimum size=0.8cm}]
                \node (G_1) at (0.0, 0.0) {$G_1$};
                \node (T_2) at (1.5, 0.8) {$T_2$};
                \node (G_2) at (3.0, 0.0) {$G_2$};
                \node (T_3) at (4.5, 0.8) {$T_3$};
                \node (G_3) at (6.0, 0.0) {$G_3$};
            \end{scope}
            \begin{scope}[every edge/.style={->, thick, draw}]
                \path [bend left=10] (G_1) edge node {} (T_2);
                \path [bend right=10] (T_2) edge node {} (G_2);
                \path [bend left=10] (G_2) edge node {} (T_3);
                \path [bend right=10] (T_3) edge node {} (G_3);
                \path [bend right=90] (T_2) edge node {} (G_3);
            \end{scope}
        \end{tikzpicture}
    }
    \caption{A gene regulatory network (GRN). $G_i$ represents a gene and $T_i$ represents a transcription factor. $E_i$ is the error applied to the corresponding node. Endogenous nodes are shown in big circles, exogenous nodes are shown in small narrow circles, and extraneous nodes are shown in dashed circles. \subref{fig:grn-causation-a} The actual GRN of four genes $G_1, ..., G_4$ with only three genes expressed in natural situations. \subref{fig:grn-causation-b} The network infered through experimentation.}
    \label{fig:grn-causation}
\end{figure}

Assume that the experimenter tries to discover the causal relationship between genes using interventions. He/She constructs an incomplete model $\calM_\text{G} = \langle \mathbfit{X}, \mathbfit{E}, \calH, \mathbff, \mathbbP_\calbfE \rangle$ with $\mathbfit{X} = \langle G_1, T_2, G_2, T_3, G_3 \rangle$ and $\mathbfit{E} = \langle E_0, ..., E_5 \rangle$. Then she uses Pearl's rules of intervention. When she wants to see whether there is a direct causal relationship between $T_2$ and $G_3$, she applies two interventions $\langle G_1, T_2, G_2, T_3 \rangle \dooeq \langle 0, 0, 0, 1 \rangle$ and $\langle G_1, T_2, G_2, T_3 \rangle \dooeq \langle 0, 1, 0, 1 \rangle$. These interventions result in two different probability measures $G_3 \sim \text{Normal}(0, \sigma_1)$ and $G_3 \sim \text{Normal}(1, \sigma_2)$ and a direct causal relation is inferred. Figure \ref{fig:grn-causation} \subref{fig:grn-causation-b} shows the inferred network.

The inferred network is correct -- indeed, $T_2$ is a direct cause of $G_3$ with respect to $\calM_\text{G}$. The problem is that there are cases that the experimenter is interested in how the system naturally behaves, while what he/she infers through interventions tells him/her how the system works under interventions. In the natural situation $T_2$ has no causal influence on $G_3$.

The front-door interventional interference can be explained as the enforcement of a generic outcome to attain a measure zero value. In other words, the intervention has caused the situation that does not naturally occur.

\subsection{Back-Door Cross-Domain Invalidity}
Similar to the front-door cross-domain invalidity, the back-door cross-domain invalidity leads to the false detection of spurious causal relationships. Unlike front-door cross-domain invalidity, it is caused by a back-door interventional interference. A back-door interventional interference is the inference that finds its way through the path from source of intervention to the potential cause. For example, if an intervention is conducted to test the causal relationship from $X_i$ to $X_j$, then if the intervention makes an invisible causal path to $X_j$ that is not part of the model, then a back-door interventional interference has taken place.

In the gene regulatory network hypothetical experiment that we discussed in the previous sub-section, a possible scenario that might cause back-door interventional interference, is that the experimenter applies a protein that is supposed to make a protein complex with $T_2$ and activate $G_2$, but the protein complex itself interferes in the regulation of $G_3$ through a secondary chemical path. If he/she intends to test the direct causal relationship between $T_2$ and $G_3$ using Pearl's second rule of intervention, then he/she might end up with a false direct link from $T_2$  to $G_3$.

Both front-door and back-door cross-domain invalidates, can be considered a sort of external invalidity; They happen when the resulted model in one context is not valid in a second context. 

There is also a simpler way to interpret these invalidates. Imagine that an experimenter tries to test the causal effect of $X_i$ on $X_j$. The interventional interference can be thought of as the interference in the measure of the exogenous generic outcome that is the direct cause of the the endogenous generic outcome $X_j$.

\section{Final Remarks}
\label{sec:final-remarks}
In this section, we will bring a short discussion on how \textit{a priori} knowledge may appear in the network. We will then conclude with a conjuncture on how the current causal discovery algorithms can be improved.

\subsection{Distribution of A Priori Information in SCMs}
As we discussed in \ref{sec:distinction}, one of the interesting features of SCMs is that they can have \textit{a priori} nodes. However, we do not insist that \textit{a priori} knowledge only appears in the form of isolated relations. 

In subsection \ref{sec:marginalization}, we brought an example where marginalizing an \textit{a priori} node formed a successor node that was not \textit{a priori}. This suggests that the information that appear in the structural equation of a node may be a combination of \textit{a priori} and \textit{a posteriori} information. Even if we re-configure the structure of an \textit{a priori} module by altering some of its edges, the \textit{a priori} node of that module may no longer be \textit{a priori}. In appendix \ref{sec:reversibility}, we bring an example where the re-configuration of the \textit{a priori} module of an SCM has re-distributed the information among other nodes of the network, and also changed some \textit{a posteriori} node to become \textit{a priori}. This means that not only \textit{a priori} information may not be atomic, they may also appear in other forms than isolated structural equations.

Such information that is truly \textit{a priori} may be scarce or even impossible. But there exist information that are gained prior to the intervention or observation of the system. These information may improve the process of causal inference in causal discovery algorithms. We will discuss the usage of this type of information in the remainder of this section.

\subsection{Possible Improvement of Causal Discovery Algorithms using Prior-To-Experience Knowledge}
There are many algorithms for causal discovery that have been developed around he idea of interventions. These algorithms can be roughly divided into the categories of the algorithms that use experimentation, those that use purely observational data, and the hybrid ones, that use a combination of the two. The experimentation-based methods perform randomized controlled trials to generate the network; The methods that are based on purely observational data either use constraints based on the underlying causal assumptions to limit the number of possible networks (LCD \cite{cooper1997simple}, Y-Structures \cite{mani2012theoretical}, PC \cite{spirtes2000causation}, IC \cite{geiger1990identifying}), or score networks based on criteria like likelihood and complexity (Bayesian Network Learning \cite{heckerman1995learning}, LiNGAM \cite{shimizu2014lingam}); And the hybrid methods combine the information gained from both observation and experimentation (JCI \cite{mooij2018joint}).

All of these methods complement the traditional regime used in science; the hypothetico-deductive method. The hypothetico-deductive method has been in play for centuries. Unlike these algorithms, the hypothetico-deductive method is not fully dependent on data. A scientist can make predictions about the causal structure of behavior of a system without having done a single experiment. In some cases he/she doesn't even need to observe the system to predict its structure.

Based on the reductionist model of science, one of the major reasons that the scientist can outperform a causal discovery algorithm may be his/her ability to construct her theoretical causal network upon well-founded knowledge of the underlying mechanism of the system. After all, it is much easier to predict that ``smoking is a cause of cancer'' if we know that ``Nitrosamines is a cause of mutogenesis''. 

The scientist's knowledge is prior to experiment and the aforementioned algorithms do not explicitly use it in the construction of SCMs. It might be case that this knowledge presents itself in the form of some edges of the network. In this case, these edges are formed before data is fed to the algorithm and cast a constraint upon the inference of the network. This might also be the case that this knowledge appears in the network in a holistic form (e.g. in two structural equations of the network). 

To the best of our knowledge, to this day, the only effort to exploit this type knowledge in algorithmic causal inference has been done in \citen{cox2019improving}.

\bibliography{references}

\phantomsection
\section*{Appendices}
\addcontentsline{toc}{section}{Appendices}
\setcounter{subsection}{0}
\renewcommand{\thesubsection}{\Alph{subsection}}

\subsection{On the Suitability of Using the New Terminology in Probabilistic Causality}
\label{sec:terminology}
Random variables are measurable functions that map a sample space $\calX$ to the set of real numbers with respect to a given \textsigma-algebra $\calF_\calX$. By definition, $X : \calX \to \mathbbR$ is called a random variable if it is $\calF_\calX$-measurable \cite{spreij2012measure}. In this paper, we used the term generic outcome in place of random variables. A generic outcome is an outcome --possibly but not necessarily in the physical world-- with respect to a measurable space, without any possible value bound to it. A generic outcome $X : \calX$ does not define a mapping like a random variable $X : \calX \to \mathbbR$ does. Indeed, a random variable can map the domain of a generic outcome to the domain of another generic outcome. In this appendix, we will bring a series of arguments on why we defined and used the proposed terminology extensively throughout this paper.

\subsubsection{Random variables are mappings; generic outcomes are not}
A convenient aspect of generic outcomes is that they are not functions. In fact, a generic outcome can be directly seen as the representative of a physical phenomenon. A generic outcome can also be a hypothetical (or purely mathematical) object. However, random variables are mappings by definition. If an experimenter wants to measure the the joint probability of two distinguishable phenomena in the physical world using random variables, he/she has to map these outcomes to a new space, measure their joint probability, and map them back to the original space in order to interpret them and gain an understanding of what happens in the actual world.

Indeed, in order to measure the probability of a physical phenomenon, one has to measure the pre-image of the assigned random variable. Instead, he/she could measure the phenomenon directly. A generic outcome makes it possible to do the direct measuring more easily.

\subsubsection{Random variables have to be real-valued; generic outcomes do not}
A major problem that we encountered using random variables is that they are not defined on desirable spaces. For example, if someone wants to work with the gender of human as a random variable, he/she simply fails because of the mere definition of random variables. A random variable can not take the values ``male'' or ``female''. This problem is especially prominent when working with mathematics that have physical aspects; in a causal network, nodes generally bear physical meaning and that meaning is not necessarily compatible with what the set of real numbers can show.

Some articles may refer to random variables with co-domains defined on arbitrary set as random elements \cite{frechet1948elements}. Some of them neglect the exact mathematical definition of random variables and define them on any desirable measurable co-domains. While the usage of random element as a successor of random variable, or freely defining random variables on any co-domain, remedies this problem, the other problems with random variables are not covered by random elements or general co-domains.

\subsubsection{Random variables are redundant in the definition of SCMs}
A random variable can be roughly seen as a generic outcome coupled with a measurable function. However, some definitions of SCMs decouple the function and use it separately, because an exogenous variable does not need to be a function. Mooij \textit{et. al} (among others) have avoided using random variables and used measurable functions unitedly with measurable spaces instead \cite{mooij2016joint}. Their definition of SCM contains a measurable space in place of each exogenous variable and a measurable space coupled with a measurable function in place of each endogenous variable. If they wanted to use random variables, we assume that they would end up defining extra unnecessary spaces for exogenous variables. They instead avoided random variables completely.

An SCM does not need random variables to be defined. However, a measurable space is itself a purely mathematical object. A measurable space $\langle\mathbbR, \calB(\mathbbR)\rangle$ can be the speed of an airplane or the wavelength of the color of a car. A generic outcome on the other hand is not devoid of physical sense. A generic outcome can be observed, but a measurable space is not observable. It is nonsensical to intervene a space, yet an outcome can be intervened.

Random variables are not a necessary part of the discussion around probabilistic causality. If nothing, they will only add a layer of complexity to the mathematics of SCMs.

\subsubsection{The proposed taxonomy is in harmony with Rubin's causality regime}
Potential outcome is a familiar name in the counterfactual regime of causality \cite{hernan2020causal}. In our taxonomy presented in section \ref{sec:basic-notation}, we used the notion a potential outcome in nearly the same sense. 

According to Hern{\'a}n and Robins, ``potential outcomes'' or ``counterfactual outcomes'' are outcomes that, depending on the situation, may or may not occur (that is, they may be counter to the fact) \cite{hernan2020causal}. What we mean by a potential outcome is the specific value that a generic outcome can attain regardless of it having just happened in the physical world.

The taxonomy that we proposed reconciles the meaning of potential outcomes, actual outcomes, and interventions. The distinction between potential and actual outcome are not significant when the mathematical aspect of probabilistic causality is considered, especially when either of Pearl's or Rubin's causal regimes are considered in isolation. However, if the causal meaning is to be given to the mathematical model, these concepts can be helpful.

\subsection{Remodeling of SCMs with A Priori Modules}
\label{sec:reversibility}
An \textit{a priori} relationship is expected to be symmetric in terms of causation. Hence, the general use of directional edges to declare causality is not applied to \textit{a priori} relations in an SCM. In principle, the direction of edges in an \textit{a priori} is not determined by causation. For example, if $X_1 = f_1(X_2, X_3)$ is an \textit{a priori} relation, then --unlike a usual relation-- neither $X_2$ nor $X_3$ is a direct cause of $X_1$. This gives some freedom on adding or removing some edges in a network with an \textit{a priori} relation while preserving essential properties of the network. To analyze which edges can be added or removed, we define modules that encompass \textit{a priori} nodes. We refer to these modules as ``\textit{a priori} modules''. If edges are added to or removed from the network with regards to the \textit{a priori} modules and without violating the causal relations in that network, we say that the module has been re-configured and call the resulting module a configuration of that module. We call the process of re-configuration of an SCM with \textit{a priori} modules, the \textit{remodeling} of that SCM. In essence, the remodeling of an SCM $\calM$ gives a second SCM $\calM'$ that has all of the causal properties of $\calM$. We call $\calM'$ a causally equivalent SCM of $\calM$ and call the collection of such SCMs the causal-equivalence class of $\calM$. In the remainder of this appendix, we will discuss the causally equivalent SCMs of a generic acyclic SCM with a single \textit{a priori} module.

\begin{figure}
    \centering
    \subcaptionbox{\label{fig:a-priori-module-a}}{
        \begin{tikzpicture}
            \node at (-3.8, 0.0) {};
            \node at (3.8, 0.0) {};
            \draw [fill=lightgray, opacity=0.5] plot [smooth cycle] coordinates {(0, 2.0) (-0.7, 1.7) (-1.8, 1.1) (-1.2, 0.2) (0.0, -0.6) (1.2, 0.2) (1.8, 1.1) (0.7, 1.7)};
            \begin{scope}[every node/.style={circle,thick,draw,minimum size=0.8cm}]
                \node (X) at (0.0, 0.0) {$X$};
                \node (Y) at (-1.1, 0.9) {$Y$};
                \node (Z) at (1.1, 0.9) {$Z$};
                \node (A) at (-2.3, 1.5) {$A$};
                \node (B) at (2.3, 1.5) {$B$};
                \node (C) at (0.0, -1.3) {$C$};
            \end{scope}
            \begin{scope}[every node/.style={circle,draw,inner sep=0,minimum size=0.5cm}]
                \node (E_A) at (-2.9, 2.5) {$E_A$};
                \node (E_B) at (2.9, 2.5) {$E_B$};
                \node (E_X) at (0, 1.5) {$E_X$};
                \node (E_Z) at (1.4, 2.2) {$E_Z$};
            \end{scope}
            \begin{scope}[every edge/.style={->, thick, draw}]
                \path (X) edge node {} (C);
                \path (Y) edge node {} (X);
                \path (Z) edge node {} (X);
                \path (E_X) edge node {} (X);
                \path (A) edge node {} (Y);
                \path (B) edge node {} (Z);
                \path (E_A) edge node {} (A);
                \path (E_B) edge node {} (B);
                \path (E_Z) edge node {} (Z);
            \end{scope}
        \end{tikzpicture}
    }
    \subcaptionbox{\label{fig:a-priori-module-b}}{
        \begin{tikzpicture}
            \node at (-3.8, 0.0) {};
            \node at (3.8, 0.0) {};
            \draw [opacity=0.5, dashed] plot [smooth cycle] coordinates {(-0.1, 1.7) (-0.4, 2.5) (-1.2, 2.6) (-1.2, 1.9) (-0.7, 1.2) (-2.2, 0.1) (-2.2, -1.3) (-1.0, -1.3) (0, -0.8) (1.0, -1.3) (2.2, -1.3) (2.2, 0.1) (+0.7, 1.2)};
            
            \draw [fill=lightgray, opacity=0.5] plot [smooth cycle] coordinates {(-0.5, 1.0) (-0.5, 0.0) (1.0, -1.1) (2.0, -1.1) (2.0, -0.1) (+0.5, 1.0)};
            \draw [fill=lightgray, opacity=0.5] plot [smooth cycle] coordinates {(0.5, 1.0) (0.5, 0.0) (-1.0, -1.1) (-2.0, -1.1) (-2.0, -0.1) (-0.5, 1.0)};
            
            \begin{scope}[every node/.style={circle,thick,draw,minimum size=0.8cm}]
                \node (X) at (0.0, 0.5) {$X$};
                \node (A) at (-1.8, 1.3) {$A$};
                \node (B) at (1.8, 1.3) {$B$};
                \node (Y) at (-1.5, -0.6) {$Y$};
                \node (Z) at (1.5, -0.6) {$Z$};
                \node (C) at (0.0, -1.3) {$C$};
            \end{scope}
            \begin{scope}[every node/.style={circle,draw,inner sep=0,minimum size=0.5cm}]
                \node (E_A) at (-2.9, 2.5) {$E_A$};
                \node (E_B) at (2.9, 2.5) {$E_B$};
                \node (E_X) at (-0.8, 2.2) {$E_X$};
                \node (E_Z) at (+0.8, 2.2) {$E_Z$};
            \end{scope}
            \begin{scope}[every edge/.style={->, thick, draw}]
                \path (X) edge node {} (C);
                \path (X) edge node {} (Y);
                \path (X) edge node {} (Z);
                \path (E_X) edge node {} (X);
                \path (E_Z) edge node {} (X);
                \path (A) edge node {} (X);
                \path (B) edge node {} (X);
                \path (E_A) edge node {} (A);
                \path (E_B) edge node {} (B);
            \end{scope}
        \end{tikzpicture}
    }
    
    \caption{Schematic of an SCM with an \textit{a priori} module whose generic outcomes are related to the experiment of throwing a dice and its causally equivalent network. \subref{fig:a-priori-module-a} The original network where $X$ is an \textit{a priori} node. \subref{fig:a-priori-module-b} A causally equivalent network of \subref{fig:a-priori-module-a} with two \textit{a priori} modules. The dashed area indicates the nodes inside the original network. Each gray area is an \textit{a priori} module.}
    \label{fig:a-priori-module}
\end{figure}
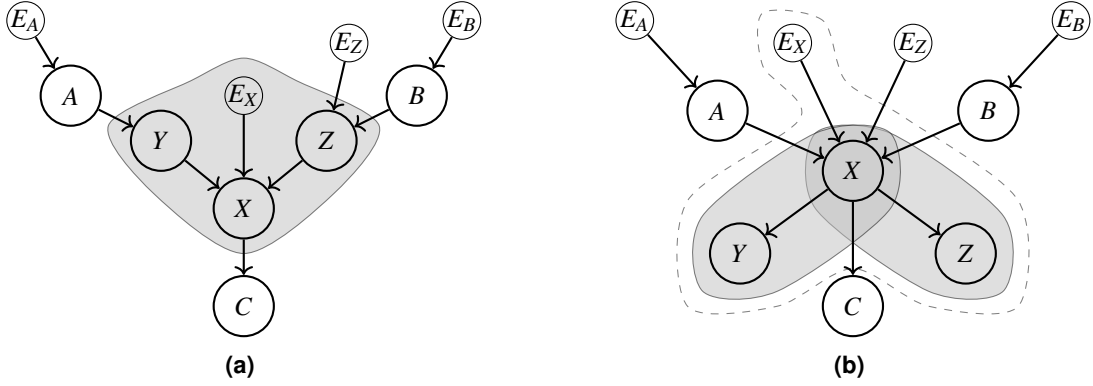

In an acyclic SCM $\calM = \langle \mathbfit{X}, \mathbfit{E}, \calH, \mathbff, \mathbbP_\calbfE \rangle$, we denote the \emph{\textit{a priori} module} of an \textit{a priori} node $X_i \in \mathbfit{X}$ using the function $\ap_\calM(X_i)$ and define it as a function that returns the set of nodes that is constructed using the following axioms:
\begin{enumerate}[label={(\roman*)}]
    \item $X_i \in \ap_\calM(X_i)$
    \item $\forall X_j \in \mathbfit{X} : X_j \in \ap_\calM(X_i) \text{ and } X_j$ is \textit{a priori} $\Rightarrow \pa_\calH(X_j) \subseteq \ap_\calM(X_i)$
\end{enumerate}
Figure \ref{fig:a-priori-module} \subref{fig:a-priori-module-a} shows a network with a single \textit{a priori} node $X$. In this SCM, $\ap(X) = \{X, Y, Z, E_X\}$.

Let $\mathbfit{V}_{\calI'} \subseteq \calV_\calH$ be the \textit{a priori} module of the only \textit{a priori} node $X_i \in \mathbfit{X}$ in an acyclic SCM $\calM = \langle \mathbfit{X}, \mathbfit{E}, \calH, \mathbff, \mathbbP_\calbfE \rangle$. In relation to $\mathbfit{V}_{\calI'}$, four types of edges in $\calH$ can be distinguished: 
\begin{enumerate}[label={(\roman*)}]
    \item \emph{interior edges} of $\mathbfit{V}_{\calI'}$, formally defined as $\left\{\left(V_j, V_k\right) \in \calD_\calH \mid V_j \in \ap_\calM(X_i) \text{ and } V_k \in \ap_\calM(X_i) \right\}$,
    \item \emph{exterior edges} of $\mathbfit{V}_{\calI'}$, formally defined as $\left\{\left(V_j, V_k\right) \in \calD_\calH \mid V_j \not\in \ap_\calM(X_i) \text{ and } V_k \not\in \ap_\calM(X_i) \right\}$,
    \item \emph{incoming edges} of $\mathbfit{V}_{\calI'}$, formally defined as $\left\{\left(V_j, V_k\right) \in \calD_\calH \mid V_j \not\in \ap_\calM(X_i) \text{ and } V_k \in \ap_\calM(X_i) \right\}$,
    \item and \emph{outgoing edges} of $\mathbfit{V}_{\calI'}$, formally defined as $\left\{\left(V_j, V_k\right) \in \calD_\calH \mid V_j \in \ap_\calM(X_i) \text{ and } V_k \not\in \ap_\calM(X_i) \right\}$.
\end{enumerate}
When remodeling an SCM, exterior edges can not be added or removed, because if $\left(V_j, V_k\right)$ is an exterior edge, it means that $V_j$ is a direct cause of $V_k$. Incoming and outgoing edges may be added or removed -- albeit with some restrictions. Only parents and children of the \textit{a priori} module can and must form new edges crossing the module. Interior edges may also be added or removed, but their distributions may not change. Intuitively, the re-configuration must not demolish any pre-existing causal path and must not form any new one. This leads to the formal definition of causal equivalence. This definition can be extended to cover more general class of SCMs, but this requires further considerations.

Let $\calM = \langle \mathbfit{X}, \mathbfit{E}, \calH, \mathbff, \mathbbP_\calbfE \rangle$ be an acyclic SCM with a single \textit{a priori} module $\mathbfit{V}_{\calI'} = \ap_\calM(X_i)$. We call $\calM' = \langle \mathbfit{X}, \mathbfit{E}, \calH', \mathbff', \mathbbP_\calbfE \rangle$ a causally equivalent SCM of $\calM$, if and only if:
\begin{enumerate}[label={(\roman*)}]
    \item for each $V_j, V_k \in \calV_\calH \setminus \mathbfit{V}_{\calI'}$:
    \begin{enumerate}[topsep=0pt, label={(\alph*)}]
        \item $\left(V_j, V_k\right) \in \calD_\calH \Leftrightarrow \left(V_j, V_k\right) \in \calD_{\calH'}$,
        \item $\exists V_l \in \mathbfit{V}_{\calI'} : V_j \in \an_\calH(V_l) \text{ and } V_k \in \de_\calH(V_l) \Leftrightarrow \exists V_{l'} \in \mathbfit{V}_{\calI'} : V_j \in \an_{\calH'}(V_{l'}) \text{ and } V_k \in \de_{\calH'}(V_{l'})$,
    \end{enumerate}
    \item for each $V_j \in \calV_\calH \setminus \mathbfit{V}_{\calI'}$:
    \begin{enumerate}[topsep=0pt, label={(\alph*)}]
        \item $\exists V_k \in \mathbfit{V}_{\calI'} : \left(V_j, V_k\right) \in \calD_\calH \Leftrightarrow \exists V_{k'} \in \mathbfit{V}_{\calI'} : \left(V_j, V_{k'}\right) \in \calD_{\calH'}$,
        \item $\exists V_k \in \mathbfit{V}_{\calI'} : \left(V_k, V_j\right) \in \calD_\calH \Leftrightarrow \exists V_{k'} \in \mathbfit{V}_{\calI'} : \left(V_{k'}, V_j\right) \in \calD_{\calH'}$,
    \end{enumerate}
    \item and for $\mathbbP_\calE$-almost every $\bm{e} \in \calbfE$ for every $\bm{x} \in \calbfX$, $f'(\bm{e}, \bm{x}) = f(\bm{e}, \bm{x})$.
\end{enumerate}
Intuitively, criterion (i) guarantees the causal paths between two nodes that are outside module $\mathbfit{V}_{\calI'}$ are preserved, criterion (ii) guarantees the causal paths that pass through $\mathbfit{V}_{\calI'}$ are preserved, and criterion (iii) guarantees the equality of the measure of all observed values in both $\calM$ and $\calM'$.

In the network shown in figure \ref{fig:a-priori-module}, $X$, $Y$ and $Y$ can be the generic outcomes in the dice-throwing example in section \ref{sec:distinction} forming an \textit{a priori} module. If the generic outcomes outside this module (i.e. $A$, $B$, $C$, $E_A$ and $E_C$) are arbitrary, then we will have a system of structural equations as folows:
\begin{equation}
    \begin{cases}
        A = f_1(E_A) = E_A\\
        B = f_2(E_B) = E_B\\
        Y = f_3(A) = \begin{cases}
            \text{odd} & A = 0\\
            \text{even} & A = 1\\
        \end{cases}\\
        Z = f_4(B + E_Z) = \begin{cases}
            \leq 3 & B + E_Z \leq 2\\
            > 3 & B + E_Z > 2\\
        \end{cases}\\
        X = f_5(Y, Z, E_X) = \begin{cases}
            1 & Y = \text{odd} \text{ and } Z \leq 3 \text{ and } E_X = 0\\
            2 & Y = \text{even} \text{ and } Z \leq 3 \text{ and } E_X = 1\\
            3 & Y = \text{odd} \text{ and } Z \leq 3 \text{ and } E_X = 0\\
            4 & Y = \text{even} \text{ and } Z > 3 \text{ and } E_X = 0\\
            5 & Y = \text{odd} \text{ and } Z > 3 \text{ and } E_X = 1\\
            6 & Y = \text{even} \text{ and } Z > 3 \text{ and } E_X = 0\\
        \end{cases}\\
        C = f_6(X) = 2 \times X\\
    \end{cases}\\
    \text{,}\\
\end{equation}
with $E_A, E_B, E_Z \sim \text{Ber}(\frac{1}{2})$ and $E_X \sim \text{Ber}(\frac{1}{3})$. What we know \textit{a priori} is the relation between $X$ (the number on the dice) and its parents ($Y$, $Z$, $E_X$). Other relations and measures have been aquired \textit{a posteriori}. Among the many ways to remodel the network, one easy way is the reversal of edges. One can use the same algorithm as proposed in \citen{cheuk1997structured}. We obtained the network in figure \ref{fig:a-priori-module} \subref{fig:a-priori-module-b} by reversing edges $(Y, X)$ and $(Z, X)$. The system of structural equation of this SCM is as follows:
\begin{equation}
    \begin{cases}
        X = f_5'(Y, B, E_Z, E_X) = \begin{cases}
            1 & Y = \text{odd} \text{ and } B + E_Z \leq 2 \text{ and } E_X = 0\\
            2 & Y = \text{even} \text{ and } B + E_Z \leq 2 \text{ and } E_X = 1\\
            3 & Y = \text{odd} \text{ and } B + E_Z \leq 2 \text{ and } E_X = 0\\
            4 & Y = \text{even} \text{ and } B + E_Z > 2 \text{ and } E_X = 0\\
            5 & Y = \text{odd} \text{ and } B + E_Z > 2 \text{ and } E_X = 1\\
            6 & Y = \text{even} \text{ and } B + E_Z > 2 \text{ and } E_X = 0\\
        \end{cases}\\
        Y = f_3'(X) = \begin{cases}
            \text{odd} & X \in \{1, 3, 5\}\\
            \text{even} & X \in \{2, 4, 6\}\\
        \end{cases}\\
        Z = f_4'(X) = \begin{cases}
            \leq 3 & X \leq 3\\
            > 3 & X > 3\\
        \end{cases}\\
    \end{cases}\\
    \text{,}\\
\end{equation}
with $f_1' = f_1$, $f_2' = f_2$ and $f_6' = f_6$.

One interesting property of remodeling simple SCMs with a single \textit{a priori} module is that it is not necessarily reversible. As seen, the SCM in figure \ref{fig:a-priori-module} \subref{fig:a-priori-module-b} is a remodeled SCM of the network in figure  \ref{fig:a-priori-module} \subref{fig:a-priori-module-b}. It has formed two new \textit{a priori} relations. In other words, the relationship between an SCM and its causally equivalent SCMs is not symmetric.
\end{document}